\documentclass{article}

\usepackage{PRIMEarxiv}

\usepackage[utf8]{inputenc} 
\usepackage[T1]{fontenc}    
\usepackage{hyperref}       
\usepackage{url}            
\usepackage{booktabs}       
\usepackage{amsfonts}       
\usepackage{nicefrac}       
\usepackage{microtype}      
\usepackage{lipsum}
\usepackage{fancyhdr}       
\usepackage{graphicx}       
\graphicspath{{media/}}     
\usepackage{multirow}
\pdfoutput=1

\title{Segmentation of the Veterinary Cytological Images for Fast
Neoplastic Tumors Diagnosis
}

\begin{document}
\maketitle

\begin{center}
\textbf{Jakub Grzeszczyk$^{1}$, Michał Karwatowski$^{1,2}$, Daria Łukasik$^{1}$ \\
Maciej Wielgosz$^{1,2}$, Paweł Russek$^{1,2}$, Szymon Mazurek$^{1}$ \\
Jakub Caputa$^{1}$, Rafał Frączek$^{1,2}$, Anna Śmiech$^{3}$ \\
Ernest Jamro$^{1,2}$, Sebastian Koryciak$^{1,2}$,  Agnieszka Dąbrowska-Boruch$^{1,2}$ \\
Marcin Pietroń$^{1,2}$, Kazimierz Wiatr$^{1,2}$ \\}

\hspace{5cm}

$^{1}$ \quad ACC Cyfronet AGH, Nawojki 11, 30-950 Cracow, Poland\\
$^{2}$ \quad AGH University of Science and Technology, al. Mickiewicza 30, 30-059 Cracow, Poland\\
$^{3}$ \quad University of Life Sciences, al. Akademicka 13, 20-950 Lublin, Poland\}

\{rafalfr, mkarwat, wielgosz, russek, adabrow, jamro, pietron, koryciak, wiatr\}@agh.edu.pl \\
\{d.lukasik, j.grzeszczyk, j.caputa\}@cyfronet.pl,
\\anna.smiech@up.lublin.pl, ymmazure@cyf-kr.edu.pl

\end{center}

\hspace{5cm}

\begin{abstract}
  This paper shows the machine learning system which performs instance segmentation
of cytological images in veterinary medicine. Eleven cell types were used directly and indirectly 
in the experiments, including damaged and unrecognized categories. The deep learning models 
employed in the system achieve a high score of average precision and recall metrics, i.e. 0.94 and 0.8
respectively, for the selected three types of tumors. This variety of label types allowed us to draw a
meaningful conclusion that there are relatively few mistakes for tumor cell types. Additionally, the
model learned tumor cell features well enough to avoid misclassification mistakes of one tumor type
into another. The experiments also revealed that the quality of the results improves with the dataset
size (excluding the damaged cells). It is worth noting that all the experiments were done using a
custom dedicated dataset provided by the cooperating vet doctors.
\end{abstract}


\section{Introduction}

Nowadays, we witness massive growth of machine learning-based solutions. They are widespread in all human activity areas and facilitate technological growth. Medicine can be considered one of the up-and-coming fields when it comes to applying artificial intelligence-based methods. Annual progress in the field of AI is enormous \cite{deep2018progress}. On the flip side, there are considerable limitations when accessing human medical data necessary to develop the machine learning algorithms, as it is sensitive and law-protected information \cite{understanding2010nicholas}. It is much more convenient for researchers to reach out for the data in veterinary medicine, as it is not as strictly protected as human data. It is also worth noting that domestic animals, such as dogs, are very close to humans in their daily routines and nutritional habits, and they are mammals too. Consequently, it is easier and faster to introduce machine learning to veterinary medicine, which may help adapt it to human medicine later.

\newpage

In this paper, we address the challenge of enhancement of neoplastic tumor diagnosis using deep learning methods. The goal is to support physicians in the assessment of cytological samples. Our main contributions are as follows:
\begin{itemize}
    \item a complete ML-based engine for automation of the tumor diagnosis in vet cytology,
    \item new large dataset labeled for segmentation \cite{cyfrovetdataset},
    \item complete pipeline for segmentation in a domain of cytological images and a series of insights on how to organize the processing for the best performance,
    \item access to our source code \cite{cyfrovetsegmodel}.
\end{itemize}

 It is worth noting that the proposed elements may be considered separately and used to boost the performance of different models. In the sections to come, we present the classical approach to cytological diagnosis in veterinary and the new proposed ML-based solution.

\subsection{Standard protocol in veterinary diagnosis}
The cytology test is performed to make a fast pre-diagnosis and evaluate potential disease management. In the case of dogs, skin lesions such as tumors or pustules are often the basis for the test qualification. In our case, the sample is taken by FNB (Fine Needle Biopsy) and then transferred onto the microscope slide. In the next step, the slide is fixed and stained.  

After appropriate preparation of the diagnostic material, the physician may evaluate the slide under the microscope. 
However, the number of diagnostic spots may vary significantly, and searching the slide for diagnostically interesting spots is time-consuming. Also, the amount of time required may increase even further because it is often needed to examine many more than one slide. 

When evaluating and describing samples, it is necessary to have qualifications and adequate skills that are a domain of veterinary pathologists. The majority of doctors are not trained to examine the samples. Moreover, analyzing the images to find pathological changes in cytology samples is very laborious. Tens or sometimes hundreds of cells may be present in the sample at various diagnostic sites, and they have to be distinguished for their pathogenicity.  Consequently, the images are examined by dedicated laboratory medicine companies. This consumes the additional precious time needed for the cytological samples to be sent for examination and the results to return to the clinic, which sometimes takes weeks even. Of course, this also involves additional costs, which may be a burden for an animal owner. 

\subsection{ML-automated protocol in veterinary diagnosis}

\begin{figure}[ht]
\centering
    \includegraphics[width=0.75\textwidth]{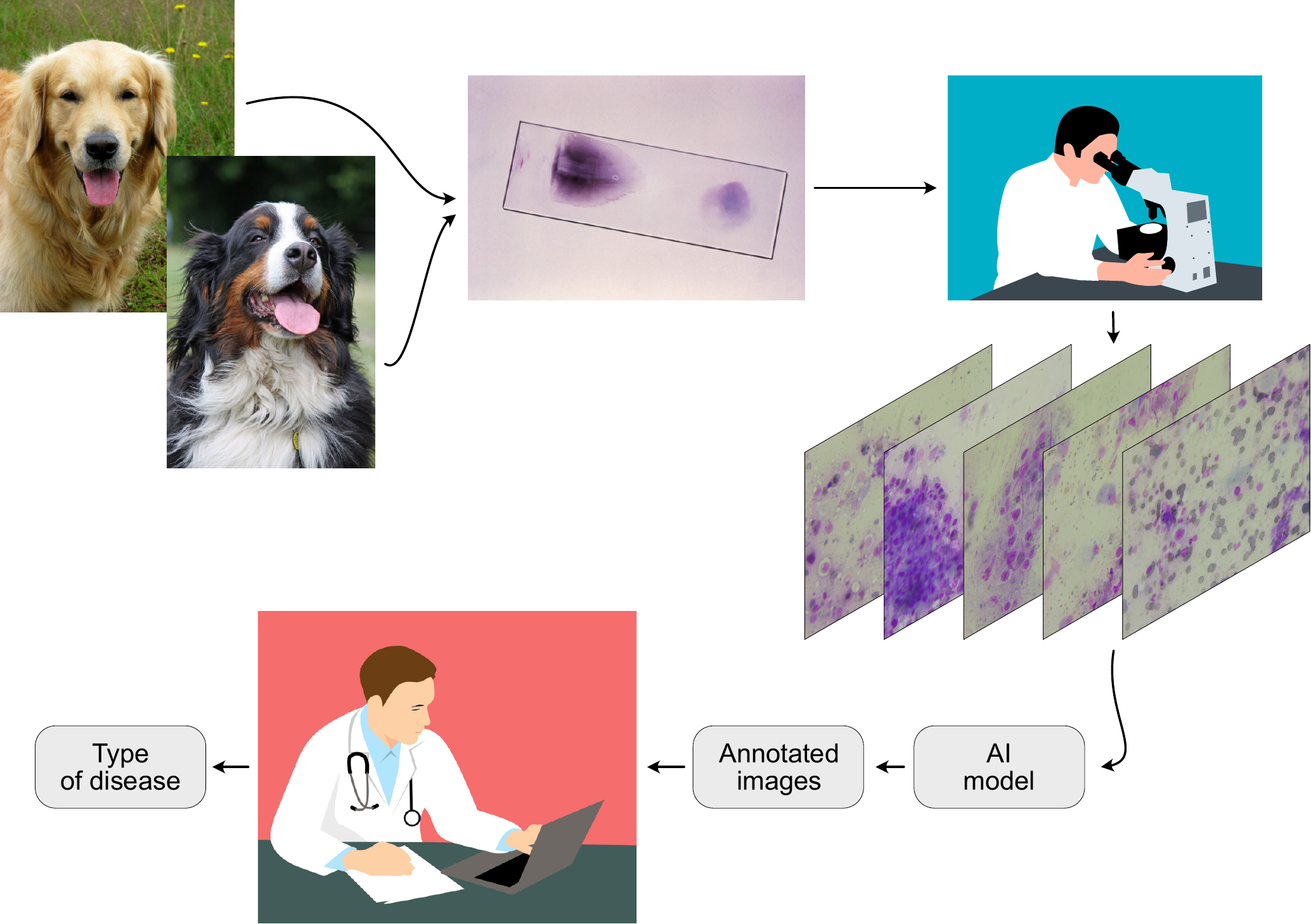}
    \caption{A diagram of the deep learning-based system for the diagnosing of pathological cancerous changes in the cytological images. Images are collected by D.Ł. (photos of the cells) and from the public domain (others).}
\label{fig:system_scheme}
\end{figure}

Today, picture digitization devices become cheaper and ubiquitous. Moreover, examining digital images and scans of the prints is more and more popular in pathology. In practice, the smartphone camera can sufficiently capture information-rich and high-quality content.
Consequently, deep learning models can support cytological diagnosis in the digital domain. Proposed by the authors, the AI-based system for diagnosing skin cancer in cytological images is presented in Figure~\ref{fig:system_scheme}.

In our scheme, the veterinarian takes pictures of the diagnostic sites of individual slides and uses ML software to check for possible tumor cells.     

Quick access to an automated tool that can reduce examination time to seconds and yield high accuracy results can sometimes be a decisive factor in therapy, especially when cancerous pathological changes are detected.

\section{Materials and Methods}

The proposed system (Fig.~\ref{fig:system_scheme})  may be perceived as a pipeline of steps to be followed to obtain the diagnosis. First, the image is acquired by the digital camera connected to the microscope. Then, the image is uploaded to the diagnosis system. In our case, the proposed system will be available online and accessed by the users via the web interface. 
Usually, it takes only several seconds to process the image and yield the results. At the current development stage of our tool, the results are segmentation maps of the image i.e. detected cells are marked.

\subsection{Collecting cytological images}
\label{samples_acquisition}
The model used in the proposed system has been trained using a custom-made database of images labeled by expert vet doctors \cite{cyfrovetdataset}. 
The final performance of a model is closely correlated with the size and quality of training data. Having that in mind, we decided to create a dataset that meets the highest standards \cite{caputa2021fast}. 

As was already briefly mentioned, cell samples were obtained from FNA (Fine Needle Aspiration) performed on dogs with skin masses. A needle with a small anuage (23-25G) is usually adequate and reduces the possibility of vessel rupture. After sampling, cellular material was spread over a microscopic slide.
The slides were prepared with the Diff-Quik staining method. Due to the short staining time, Diff-Quik is commonly used in veterinary clinics.

Next, the samples were evaluated under the DeltaOptica Evolution 300 microscope. The 40x objective has been applied, resulting in a total magnification power of 400x.
Subsequently, once a suitable spot of interest was selected by a physician, the picture was taken using the DeltaOptical DLT-Cam PRO 5 MP camera, attached to the microscope. 

When slides are digitized, the most time-consuming stage begins i.e. the cells visible in images are labeled. Our experts prepared the cell type-specific annotation suitable for instance segmentation task, as it is most closely related to activities performed by the pathologists. Creating such annotations requires tracing the boundaries of each cell and assigning it to the appropriate category. 

As the number of cells in the dataset can reach tens or hundreds of thousands, the process is time-consuming. To mitigate this, we have created a labeling tool that uses the neural network model to suggest labels for the cells. The task of the expert is to amend automatic labels. As the number of cells in the database grows, the model used by the labeling tools is also enhanced by retraining with new data. This process continues progressively.

\subsection{Dataset}\label{cha:dataset}

The complete dataset consists of 1,206 images evenly distributed in four main disease categories \cite{cyfrovetdataset}. There are three tumor categories: lymphoma, histiocytoma, mastocytoma, and inflammations. 

Lymphoma is one of the round-cell tumors. Slides are characterized by small to medium-sized cells. Nuclei are round with irregular profiles, sometimes convoluted. The cytoplasm is present in a small amount. Neoplastic mast (mastocytoma) cells contain characteristic purple intracytoplasmic granules. Nuclei, if obscured by the granules, are often hardly visible. Histiocytoma cells are round or oval. The cytoplasm is slightly basophilic. Nuclei can show a characteristic uniform shape.

For practical reasons, there are three additional cell annotations used to label cells in the dataset:
\begin{itemize}
  \item \textit{cut cell} - the cell that does not fully fit in the picture and significant parts of them are out of the border of the photo,
  \item \textit{damaged cell } - the structure of the cell is disturbed, or it is in later stages of disintegration,
  \item \textit{unrecognized cell} - though it is not damaged, cells does not show enough features required to assign them to a specific type.
\end{itemize}

Although the task is to recognize individual tumor cells, it is good to additionally diversify the samples, so the model will not over-focus on a single disease and be able to recognize additional types of cells i.e.: macrophage, eosinophil, neutrophil, lymphocyte, giant cells. From the machine learning perspective, this process may be perceived as a specific form of regularization by enriching the dataset with more data.

The variable number of cells of each type results from their natural distribution in the collected preparations. There are even more cell types present in the dataset i.e. mitotic figure, plasma cells, and fibroblast cells. However, they are not numerous enough to be included in neural network training at the moment. They were not used in the presented experiments, but annotations for them are being created to expand the model in the future (see Tab. \ref{tab:dataset}).

\begin{figure}[ht]
\centering
	\includegraphics[width=0.75\textwidth]{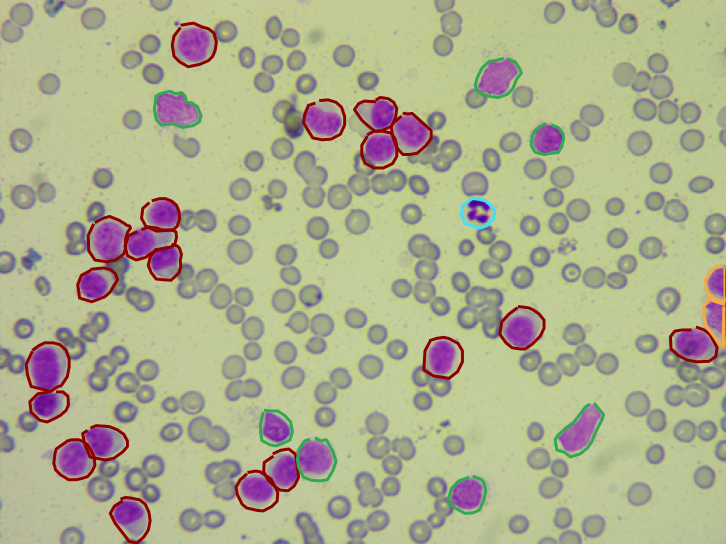}
	\caption{The annotated image lymphoma. Colors in the figure: red - lymphoma cell, orange - cut cells, green - damaged cell, aqua - neutrophil.
	\label{fig:labelled_img}}
\end{figure}

The example results of the annotations are presented in Figure~\ref{fig:labelled_img}. At the moment, the dataset includes over 70,000 annotated cells. Table ~\ref{tab:dataset} contains a summary of all cell types annotated in the dataset and their volume. Our dataset is available in the public repository \cite{cyfrovetdataset}.

\begin{table}[ht]
\centering
\caption{The cells dataset statistics}
\label{tab:dataset}
  \begin{tabular}{|l|c|c|c|c|c|}
                                 \hline
     & \textbf{Cell type} ~   & \textbf{Train count} &  \textbf{Test count} & \textbf{Val count} & \textbf{Total} \\\hline
   1. & Neoplastic mast cells  	& 9736			& 2632			& 3038			& 15406\\\hline
   2. & Histiocytoma cells	        & 4604			& 964			& 1343			& 6911\\\hline
   3. & Lymphoma cells		        & 4695			& 1253			& 1178			& 7127\\\hline
   4. & Macrophage		            & 1127			& 253			& 245			& 1625\\\hline
   5. & Eosinophil		            & 442			& 80			& 103			& 625\\\hline
   6. & Neutrophil		            & 8490			& 1904			& 1980			& 12374\\\hline
   7. & Lymphocyte		            & 717			& 171			& 192			& 1080\\\hline
   8. & Giant cell   		        & 80			& 14			& 14			&  108\\\hline
   9. & Cut		                & 5779			& 1456			& 1609			& 8844\\\hline
   10. & Unrecognized	            & 8262			& 2030			& 2281			& 12573\\\hline
   11. & Damaged		            & 7183			& 1911			& 1798			& 10893\\\hline
       & \textbf{Total}                     & 51115         & 10036         & 10743         & 71894 \\\hline
  \end{tabular}
\end{table}

\subsection{Deep Learning Models}\label{cha:models}
The models employed in the system presented in this paper deal with instance segmentation which enables the separation of individual instances of objects. This approach contrasts with classical segmentation, which creates only one global mask type for each object class. All the elements of the image at the pixel level are separated and assigned to an individual class.

We have selected two deep learning models based on Cascade Mask R-CNN \cite{CascadeRCNN} architecture. The models differ in the type of backbone used for feature extraction. One of them is the ResNeSt101 \cite{ResNeSt} which is based on skipped connections. In ResNeSt101, the input value goes in parallel with the current layer without any changes, and then it is summed with the modified input. The other model is SwinTransformer \cite{SwinTransformer} with the CBNetV2 \cite{CBNetV2} backbone architecture. It is a modification of Vision Transformer (ViT)  \cite{ViT}, which processes a single image as $16{\times}16$ pixels patches transformed later into patch vectors by a linear transformation.

\begin{figure}[htbp]
\centering
	\includegraphics[width=10 cm]{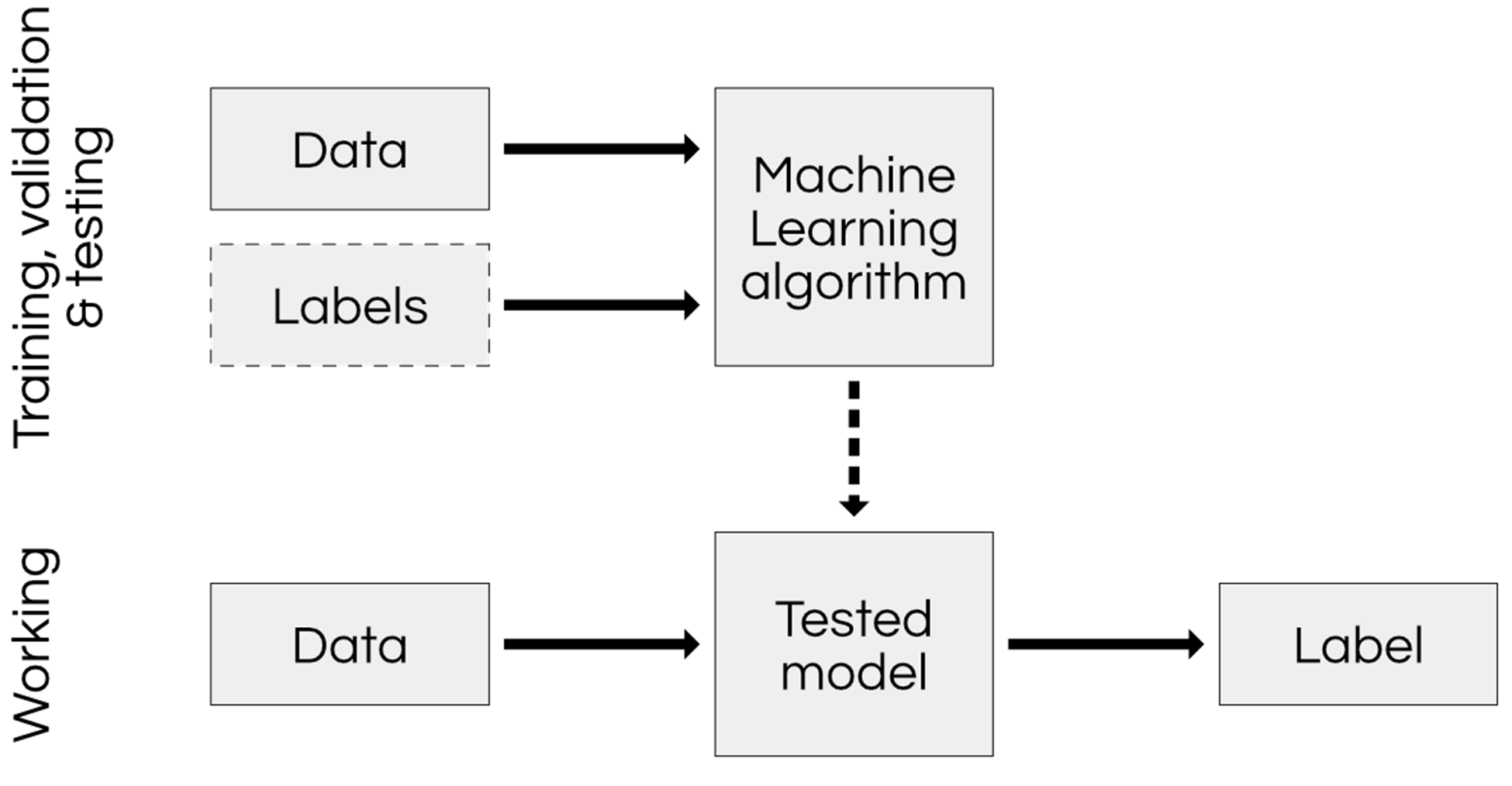}
	\caption{A diagram of the deep learning-based training and testing pipeline. \label{fig:general_pipeline}}
\end{figure}

We employ a standard machine learning training and testing pipeline presented in Figure~\ref{fig:general_pipeline}.

It requires a significant training dataset size to start training with the randomly initialized neural network parameters and train the model from scratch. Taking the limited time available for our project, such an amount of training images far exceeds the capabilities of the available expert annotators. Therefore, we initialized weights of the model with pre-trained weights for the MS COCO data set \cite{MSCOCO}. Then, we replaced the final classification heads and continued training with images from our cytological dataset. Our training parameters are presented in Table \ref{tab:training_params}. The batch size was set to two in both models, and we used resizing, random flip, normalization, and padding operations for augmentation.

\begin{table}[!ht]
\centering
\caption{Training parameters}
\label{tab:training_params}
  \begin{tabular}{|l|c|c|c|c|c|}
                                 \hline
   \textbf{Model} & \textbf{Optimizer}   & \textbf{LR} &  \textbf{LR scheduler} & \textbf{betas} & \textbf{weight decay}\\\hline
   ResNeSt101 & Adam   & 0.001  &  step & (0.9, 0.999) & 0.0001\\\hline
   SwinTransformer & AdamW   & 0.0001  &  step & (0.9, 0.999) & 0.05\\\hline
  \end{tabular}
\end{table}

The network learning process took about two and six hours on a computer server with four NVIDIA Tesla V100 graphics cards for the ResNeSt101-based and SwinTransformer-based models respectively.

\subsection{Quality assessment metrics} \label{cha:metrics}
For model performance measurement, we used metrics commonly used in object detection and recognition. 

\textit{Precision} measures how accurate the prediction is i.e. what fraction of predictions are correct. It is computed as
\begin{equation}
    \mathit{Precision} = \frac{TP}{TP + FP}~;
\end{equation}
where $TP$ and $FP$ represent true positive and false positive respectively.

\textit{Recall} measures what is the share of positive cases the model misses (a share of false negative). It is computed as
\begin{equation}
    \mathit{Recall} = \frac{TP}{TP + FN}~;
    \label{eq:rec}
\end{equation}
where $FN$ stands for false negatives. The recall is commonly referred to as \textit{sensitivity} in the medical community and it is purely a naming convention, as it is exactly the same metric. 

\textit{Specificity} is concentrated on how well the system can correctly tell negative samples from positive ones. It is computed as 
\begin{equation}
    \mathit{Specificity} = \frac{TN}{TN + FP}~;
    \label{eq:spec}
\end{equation}
where $TN$ stands for true negative.

Finally, \textit{accuracy} can be computed to measure the overall performance of the model. 
It is computed as
\begin{equation}
    \mathit{Accuracy} = \frac{TP+TN}{TP+TN+FP+FN}~.
    \label{eq:accu}
\end{equation}

\begin{figure}[ht]
    \centering
	\includegraphics[width=0.5\textwidth]{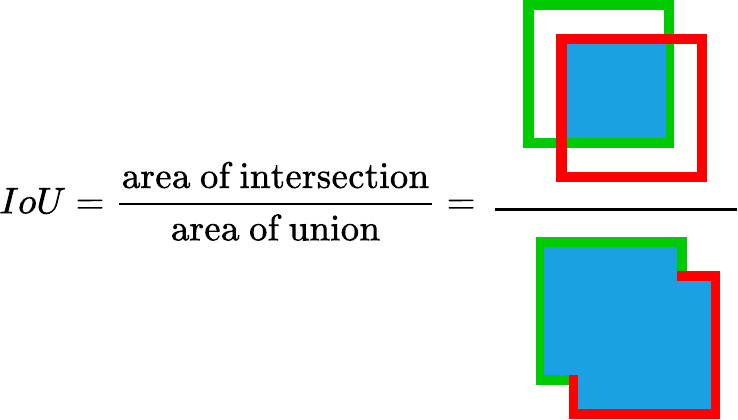}
	\caption{Intersection over Union.
	\label{fig:iou}}
\end{figure}

Specifically in image segmentation, to measure the quality of overlap of the detected object with the reference annotated by the expert, we used a metric called IoU (Fig.~\ref{fig:iou}), which stands for Intersection over Union.

For a given model, precision and recall are the competing performance metrics. Usually, the threshold value is set to distinguish the positive and negative cases. Typically, when the threshold is increased the number of false positives decreases, and the number of false negatives increases i.e. precision increases and recall decreases.

To visualize precision and recall together, we plot the precision-recall curve. It is created by varying the class probability threshold from 0.0 to 1.0, for which precision and recall are calculated. The curve allows us to select the optimal threshold for detection i.e. for best precision/recall compromise. 

Also, \textit{average precision} (AP) can be calculated to make the precision threshold independent. AP is an average precision value for the threshold range.

Both precision and recall can be calculated for different IoU thresholds. Consequently, we can tell about average precision for a given IoU value  e.g., AP@0.5 - Average precision for IoU threshold of 0.5.

To better analyze the results of our experiments, we computed a series of confusion matrices.
Confusion matrix summarises a number of $TP$, $TN$, $FP$, and $FN$ on a single chart.
This tool is widely used to visualize $TP$ and $TN$ detections present on the matrix's main diagonal and $FP$ and $FN$ outside of it. 
However, $TP$, $TN$, $FP$, and $FN$ do not cover all cases in image segmentation. The true object can be ignored by the model ("not-detected"), and the object can be detected in an empty image location ("not-present"). Therefore, we also extended the confusion matrix by adding the "not-detected" row and "not-present" column, which is helpful to control the true cells the model did not detect, and the \textit{phantom} cells it found that when the true cell was not present in the image.

The deep learning models were implemented in PyTorch using MMDetection project \cite{MMDetection}. The code for our models is available in the public repository \cite{cyfrovetsegmodel}.

\section{Related works}
Data collection and curation are one of the main challenges of developing ML-based algorithms and architecture. Available datasets are often incomplete, and some disease cases are hard to get, leading to dataset imbalance. It is hard to find a complete cytological veterinary dataset, while human datasets are much more abundant while greatly differing in labeling. A series of sample cytological datasets are presented in Table~\ref{tab:sotadatasets}.

\begin{table}[!ht]
\footnotesize
\centering
\caption{Sample datasets used in image segmentation for studying cells and tumors}
\label{tab:sotadatasets}
  \begin{tabular}{|l|c|c|c|}
                                 \hline
   \textbf{Dataset} & \textbf{Source}   & \textbf{Type} &  \textbf{Reference}\\\hline
   16 images (945 synthesized) & human & cervical cells & \cite{improved2015lu,evaluation2017lu}\\\hline
   194 images & human  &  
   \begin{tabular}{@{}c@{}}cervical cells; \textit{abnormal}  \\ and \textit{other} classes\end{tabular}
    & \cite{araujo2019deep}\\\hline
   2,540 images & human  &  
   \begin{tabular}{@{}c@{}}cervical cells; \textit{nucleus}, \textit{cluster},  \\ \textit{satellite} and \textit{background} classes\end{tabular} & \cite{atkinson2020novel}\\\hline
   1,206 images & animal &  
   \begin{tabular}{@{}c@{}}skin cells; \textit{lymphoma}, \textit{histiocytoma}  \\ and \textit{mastocytoma} classes\end{tabular} & our\\\hline
  \end{tabular}
\end{table}

Example models available for semantic segmentation for cytological images are given in Table~\ref{tab:sotamodels}. However, the majority of the models are trained for human datasets. They achieve relatively good results in terms of IoU (e.g. IoU=0.92 \cite{falk2019unet}), and most employ the U-net deep learning model. It is worth noting that a wide variety of quality assessment metrics are used for the systems since it strictly depends on the final application. Some systems are designed to yield a count of a particular cell type in the image as an output. 

\begin{table}[!tbh]
\caption{Comparison of segmentation methods. \label{tab:sotamodels}}
\begin{tabular}{|l|l|l|l|c|}
\hline
\textbf{Objective} & \textbf{Architecture} & \textbf{Type} & \textbf{Result} & \textbf{Reference} \\ \hline

\multirow{2}{*}{\begin{tabular}[c]{@{}l@{}}Various biomedical \\ segmentation problems\end{tabular}}   & \multirow{2}{*}{U-Net} 
& \begin{tabular}[c]{@{}l@{}}Glioblastom-\\ astrocytoma \\ U373 cells \\(PhC-U373)\end{tabular} & IoU=0.9203 & \cite{ronneberger2015unet} \\ \cline{3-4} 
& & \begin{tabular}[c]{@{}l@{}}HeLa cells \\ (DIC-HeLa)\end{tabular} & IoU=0.7756 & \\ \hline

\begin{tabular}[c]{@{}l@{}}Segmentation, detection and \\ classification of cell nuclei\end{tabular} & U-Net & oral cytology & IoU=0.4607 & \cite{matias2020segmentation} \\ \hline
\multirow{2}{*}{\begin{tabular}[c]{@{}l@{}}Cell nuclei segmentation\end{tabular}} & \multirow{2}{*}{\begin{tabular}[c]{@{}l@{}}CNN +\\ \\ Seeded\\ \\ watershed\end{tabular}} & \begin{tabular}[c]{@{}l@{}}benign breast cancer\end{tabular} & \begin{tabular}[c]{@{}l@{}}Hausdorff\\ distance=0.840\\ \\ Jaccard \\distance=0.776\end{tabular} & \cite{kowal2020cell} \\ \cline{3-4} & & \begin{tabular}[c]{@{}l@{}}malignant breast cancer\end{tabular} & \begin{tabular}[c]{@{}l@{}}Hausdorff\\ distance=0.781\\ \\ Jaccard \\distance=0.732\end{tabular} & \\ \hline

\multirow{2}{*}{\begin{tabular}[c]{@{}l@{}}Segmentation of \\ overlapping cells\end{tabular}} & \multirow{2}{*}{\begin{tabular}[c]{@{}l@{}}Attention\\ \\ mechanism +\\ \\ U-Net +\\ \\ Random walk\end{tabular}} & \begin{tabular}[c]{@{}l@{}}cervical cells nuclei\end{tabular} & 
$P\textsubscript{\textit{p}}=0.94 \pm 0.06$ & \\
& & & $R\textsubscript{\textit{p}}=0.95 \pm 0.05$ & \cite{zhang2020polar} \\  
& & & $Dice=0.93 \pm 0.04$ & \\ \cline{3-4}
& & \begin{tabular}[c]{@{}l@{}}cervical\\  \\ cells \\ \\ cytoplasm \\ \end{tabular} & 
$TP\textsubscript{\textit{p}}=0.94 \pm 0.06$ & \\
& & & $FP\textsubscript{\textit{p}}=0.003 \pm 0.004$ & \\
& & & $Dice=0.93 \pm0.07$ & \\ \hline
\end{tabular}
\end{table}

Different tissue preparation methods are employed, affecting the model used within the system (see  Tab.~\ref{tab:sotamodels}). A complete definition of the metrics is out of the scope of this paper, and we encourage the readers to refer to the original papers for more information.

\newpage
\section{Results}

This section presents the performance of the models described in Section \ref{cha:models} and trained on the dataset introduced in Section \ref{cha:dataset}. Additionally, we also analyze the results on a case-by-case basis. We also provide  the system application in real-world scenarios.

\subsection{Segmentation}

The performance achieved by the models is summarized in Table \ref{tab:global_metrics}. The metric scores are significantly higher for CBNetV2 architecture, which is consistent with results for the popular datasets such as COCO \cite{MSCOCO}, for example.

Results presented in Table \ref{tab:global_metrics} are averaged over all 11 cells types. Table \ref{tab:tumor_cells_metrics} comprises the same metrics specifically for distinguished tumor cells. Complete results for all cell types can be checked in Table \ref{tab:tumor_cells_metrics_extended} in Appendix \ref{appendix_A}. 

The individual tumor cell metrics values are significantly higher, showing how global results underestimate model performance in detecting and classifying certain types of cells. This finding is significant as tumor cells have the highest diagnostic value.

\begin{table}[ht]
\centering
\caption{Models' global average precision for the dataset (all cell types are analyzed together)}
\label{tab:global_metrics}
  \begin{tabular}{|l|c|c|c|}
                                 \hline
   \textbf{Metric}   & \textbf{IoU threshold} & \textbf{ResNeSt101} & \textbf{CBNetV2}\\\hline
   Average Precision & 0.50:0.95              & 0.451               & \textbf{0.510}\\\hline
   Average Precision & 0.50                   & 0.618               & \textbf{0.690}\\\hline
   Average Precision & 0.75                   & 0.570               & \textbf{0.643}\\\hline
   Average Recall    & 0.50:0.95              & 0.545               & \textbf{0.604}\\\hline
  \end{tabular}
\end{table}

\begin{table}[ht]
\centering
\caption{Segmentation metrics calculated for tumor cells categories of the test set}\label{tab:tumor_cells_metrics}
\begin{tabular}{|c|c|c|l|l|}
\hline
\multicolumn{1}{|l|}{\textbf{Category}} & \multicolumn{1}{l|}{\textbf{Metric}} & \multicolumn{1}{l|}{\textbf{IoU threshold}} & \textbf{ResNeSt101} & \textbf{CBNetV2} \\ \hline
\multirow{4}{*}{Mastocytoma}            & Average Precision                    & 0.50:0.95                                   & \textbf{0.769}      & 0.764            \\ \cline{2-5} 
                                        & Average Precision                    & 0.5                                         & \textbf{0.940}      & 0.930            \\ \cline{2-5} 
                                        & Average Precision                    & 0.75                                        & \textbf{0.940}      & 0.930            \\ \cline{2-5} 
                                        & Average Recall                       & 0.50:0.95                                   & \textbf{0.806}      & 0.799            \\ \hline
\multirow{4}{*}{Histiocytoma}           & Average Precision                    & 0.50:0.95                                   & 0.665               & \textbf{0.671}   \\ \cline{2-5} 
                                        & Average Precision                    & 0.5                                         & \textbf{0.877}      & 0.868            \\ \cline{2-5} 
                                        & Average Precision                    & 0.75                                        & 0.838               & \textbf{0.843}   \\ \cline{2-5} 
                                        & Average Recall                       & 0.50:0.95                                   & 0.740               & \textbf{0.745}   \\ \hline
\multirow{4}{*}{Lymphoma}               & Average Precision                    & 0.50:0.95                                   & 0.743               & \textbf{0.760}   \\ \cline{2-5} 
                                        & Average Precision                    & 0.5                                         & 0.955               & \textbf{0.964}   \\ \cline{2-5} 
                                        & Average Precision                    & 0.75                                        & 0.946               & \textbf{0.956}   \\ \cline{2-5} 
                                        & Average Recall                       & 0.50:0.95                                   & 0.796               & \textbf{0.812}   \\ \hline
\end{tabular}
\end{table}

\subsection{Cell count requirements analysis}

To explain discrepancies in model performance for different cell types, we need to analyze the results for individual cell types. We concentrate on CBNetV2 architecture as it achieved better results than ResNeSt101 (for ResNeSt101 results see Table \ref{tab:tumor_cells_metrics_extended} in Appendix \ref{appendix_A}). Table \ref{tab:complete_results_with_count} contains the AP@0.75 performance and number of cell instances that our CBNetV2-based model is capable to detect. AP was calculated on test split, but the intention is that results depend on the number of samples for each type in train split. 

\begin{table}[!ht]
\centering
\caption{Results for all cell types with train set abundance count\label{tab:complete_results_with_count}}
		\begin{tabular}{|l|c|c|}\hline
			\textbf{Cell type} & \textbf{AP@IoU0.75} & \textbf{train count} \\ \hline
		      Mastocytoma cell    & 0.930 & 2632 \\ \hline
                Unrecognized       & 0.385 & 2030 \\ \hline
                Damaged            & 0.557 & 1911 \\ \hline
                Neutrophil         & 0.779 & 1904 \\ \hline
                Cut                & 0.630 & 1456 \\ \hline
                Lymphoma cell      & 0.956 & 1253 \\ \hline  
                Histiocytoma cell  & 0.843 & 964  \\ \hline
                Macrophage         & 0.667 & 253  \\ \hline
                Lymphocyte         & 0.435 & 171  \\ \hline
                Eosinophil         & 0.662 & 80   \\ \hline
                Giant cell         & 0.228 & 14  \\ \hline
		\end{tabular}
\end{table}

\begin{figure}[ht]
\centering
	\includegraphics[width=0.5\textwidth]{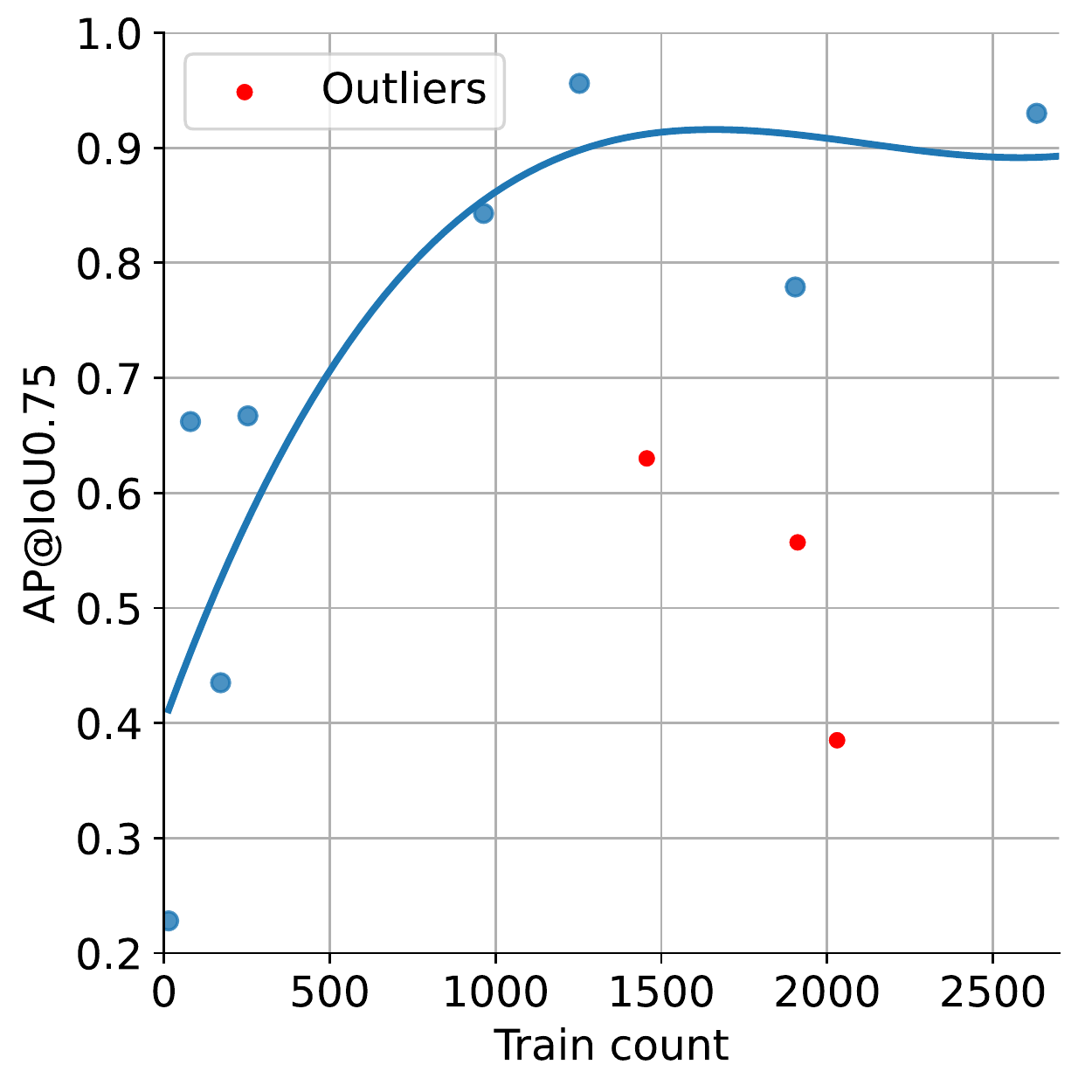}
	\caption{The trend line for AP@IoU0.75 vs. the number of cell instances in the training dataset. The outliers are marked in red. \label{fig:ap_train_plot}}
\end{figure}

We checked the correlation of the model performance with the volume of the training set for the different cell types. The trend is clearly visible on the plot in Figure~\ref{fig:ap_train_plot}. Less numerous classes have lower scores as the model does not have enough examples to learn their attributes properly. Three outlier types were not used to create the trend line: cut, damaged, or unrecognized cells.  Their corresponding results were not lower due to insufficient data but because of the higher complexity of the recognition process. Section \ref{cha:visual_analysis} provides a more in-depth analysis of the model drawbacks.

We can conclude that at least 1,000 instances of an object should be presented in the training set to achieve very good  results. Consequently, as the global AP metric value is the average of AP scores for each class, difference in metric values given in Table~\ref{tab:global_metrics} and \ref{tab:tumor_cells_metrics} can be understood. It is worth noting that fewer samples in train split also mean fewer samples in test split i.e. higher expected variance of the results.

\subsection{Confusion matrix analysis} \label{cha:conf_mtx}

The confusion matrix shows the cell categories the model often confuses and recognizes correctly. The value numbers of correctly classified cells are located on the main diagonal, and misclassifications are placed outside the diagonal. 

\begin{figure}[!htb]
\centering
    \includegraphics[width=0.75\textwidth]{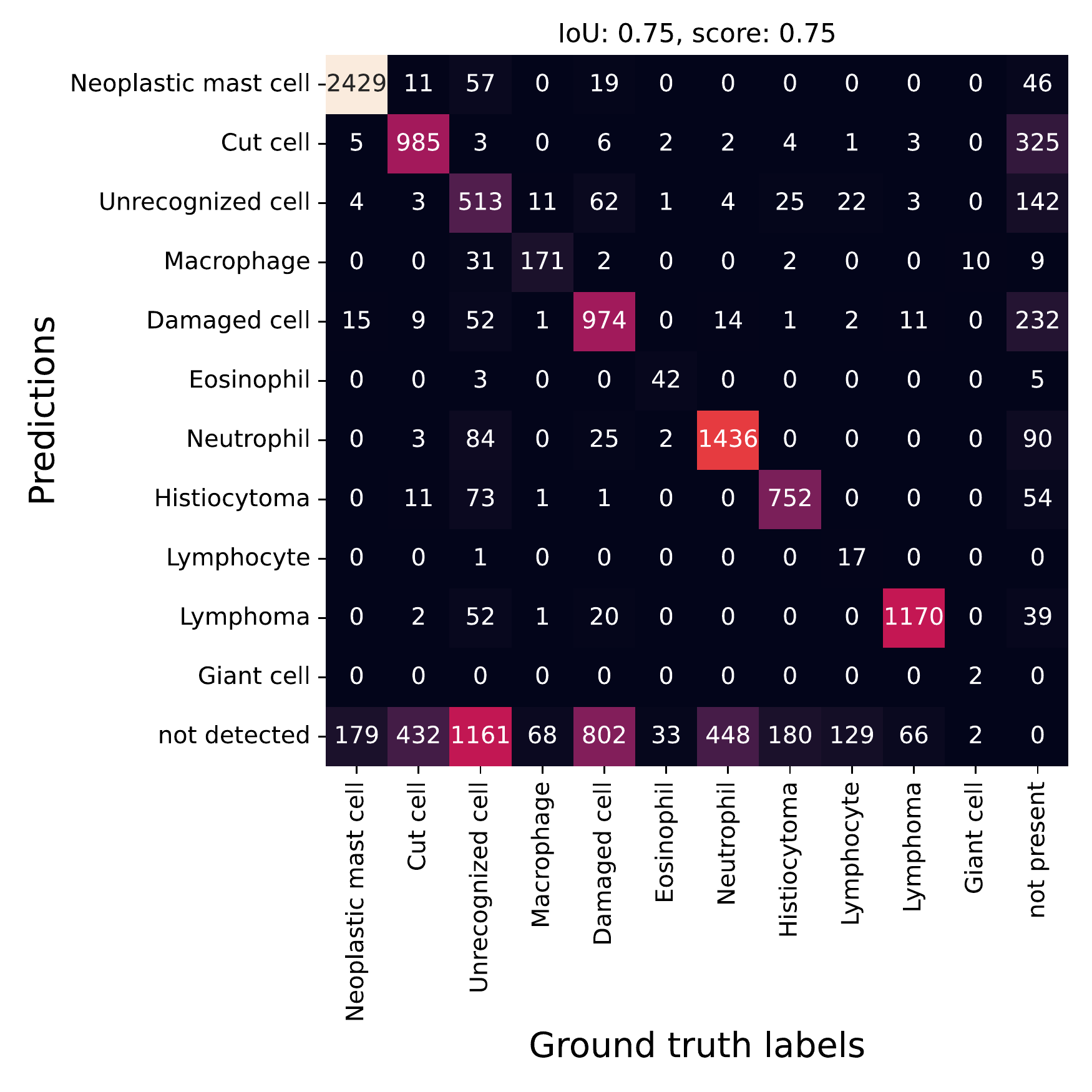}
	\caption{Extended confusion matrix for IoU=0.75 and score threshold=0.75 for CBNetV2. \label{fig:cm_swin}}
\end{figure} 

Figure \ref{fig:cm_swin} presents confusion matrix of model predictions on the test set. Correct predictions in the main diagonal agree with results from Table \ref{tab:complete_results_with_count}, but here, one can also observe which of the predicted classes were confused specifically. 

Misclassification of the three main tumor types (mastocytoma, histiocytoma, and lymphoma) regards the damaged, cut, and unrecognized classes. Those cases seem to be the most challenging, and veterinary experts can explain why. For example, the damaged or unrecognized cells predicted as lymphoma may be lymphoma in an early disintegration phase. While it still can be recognized, it is no longer diagnostically useful and should be categorized as a damaged cell. Such mistakes should be minimized in the system's further development; however, the borders between those classes are blurry and often subjective. Therefore, it may be impossible to eliminate them. There are also four inaccuracies where tumor cells were confused with macrophages. Even though it is a tiny number compared to the correct predictions (4,351), we should strive to eliminate them.

\newpage
The primary source of confusions is the \textit{not-detected} category. The main contributors to errors are unrecognized, damaged, and cut cells, adding up to $68\%$ of all not detected mistakes. Despite being reasonably numerous in the training dataset, the neutrophil is the second most challenging cell type. This type of cell is complicated to recognize for the models, and efforts should be made to improve model performance for this category.

\begin{figure}[ht]
\centering
    \includegraphics[width=0.75\textwidth]{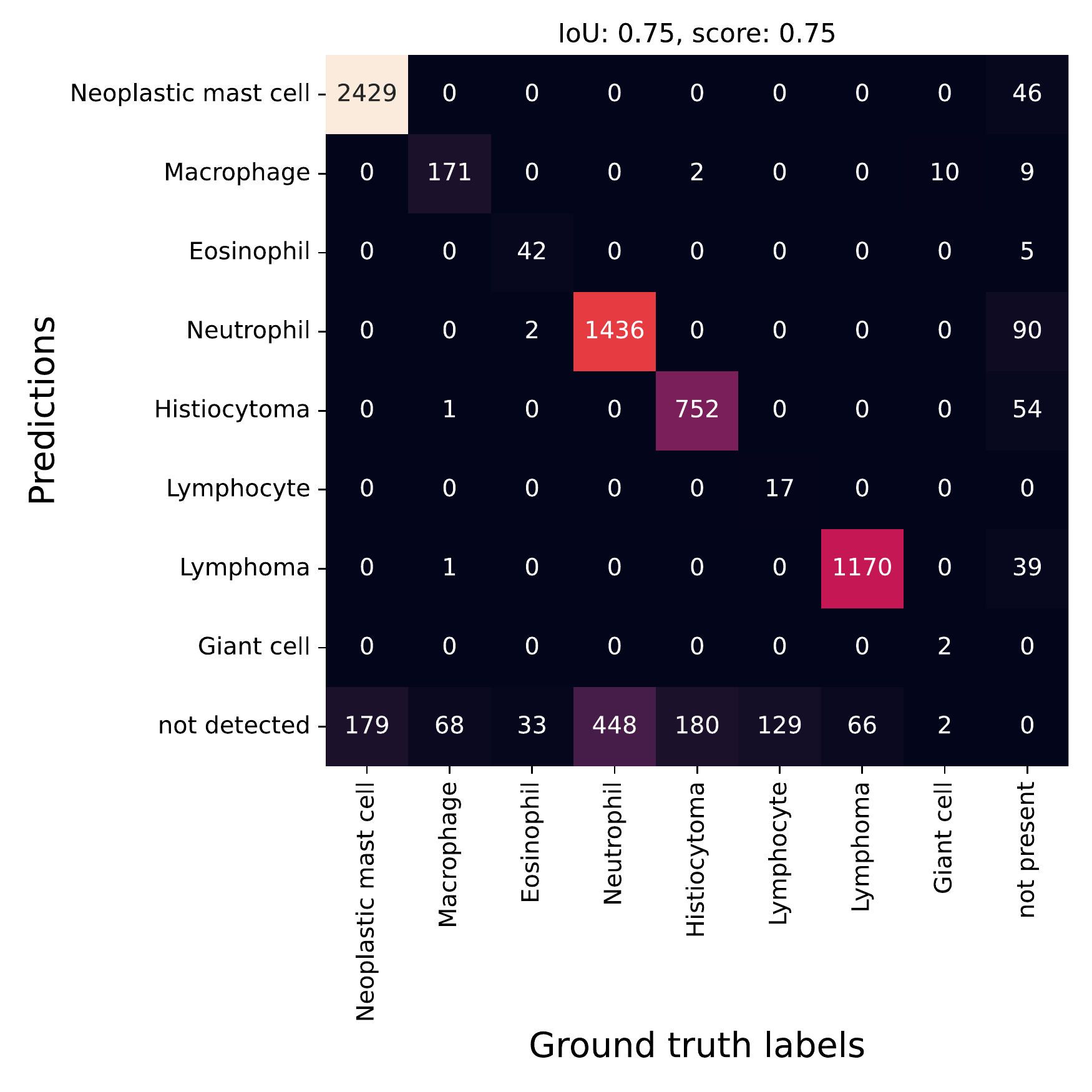}
	\caption{Basic confusion matrix for IoU=0.75 and score threshold=0.75 for CBNetV2 model.\label{fig:cm_swin_correct}}
\end{figure}

Another significant source of mistakes is the \textit{not-present} category. While all of the arguments given for \textit{not-detected} category analysis are also valid here, there is another factor. As the \textit{not-present} faults were manually verified by the expert, it was found out that a fair amount of \textit{not-present} predictions were correct in reality; the human annotator missed them, and the model pointed them correctly. That shows how challenging it is to create high-quality annotations for machine learning. This problem is further discussed in Section \ref{cha:labelong_counclusions}.

\newpage
While unrecognized, damaged, and cut cell types are helpful in model development and performance analysis, they are less interesting for a final diagnosis taken by the veterinary doctor. Those cells are mostly disregarded in disease pronouncement. For illustration, we present the same confusion matrix without those types in Figure \ref{fig:cm_swin_correct}; it is visible that mistakes between fully diagnostic cells only are much less frequent.

\begin{figure}[!ht]
\centering
	\includegraphics[width=0.75\textwidth]{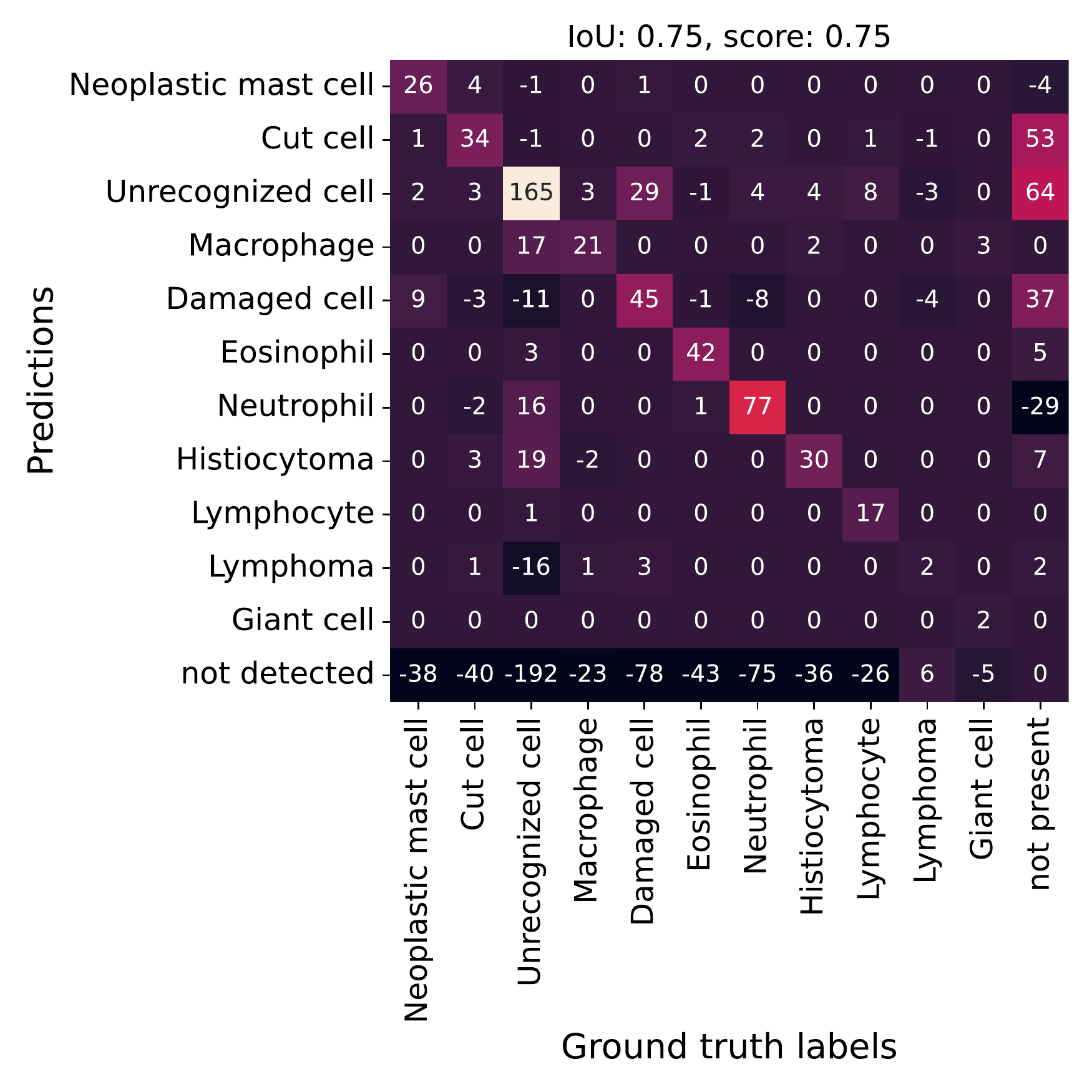}
	\caption{The difference confusion matrix for ResNeSt101 and CBNetV2.
	\label{fig:diff_cm2}}
\end{figure}

We can also compare the performance of trained models using a confusion matrix. In Figure \ref{fig:diff_cm2}, we can see the difference confusion matrix for ResNeSt101 and CBNetV2. Positive numbers on the main diagonal and negative elsewhere would indicate pure improvement. Indeed, CBNetV2 correctly found more cells in every category and, in general, detected more cells as indicated by negative numbers in \textit{not-detected} category.

\newpage
\subsection{Parameters tuning}

\begin{figure}[hp]
\centering
	\includegraphics[width=0.7\textwidth]{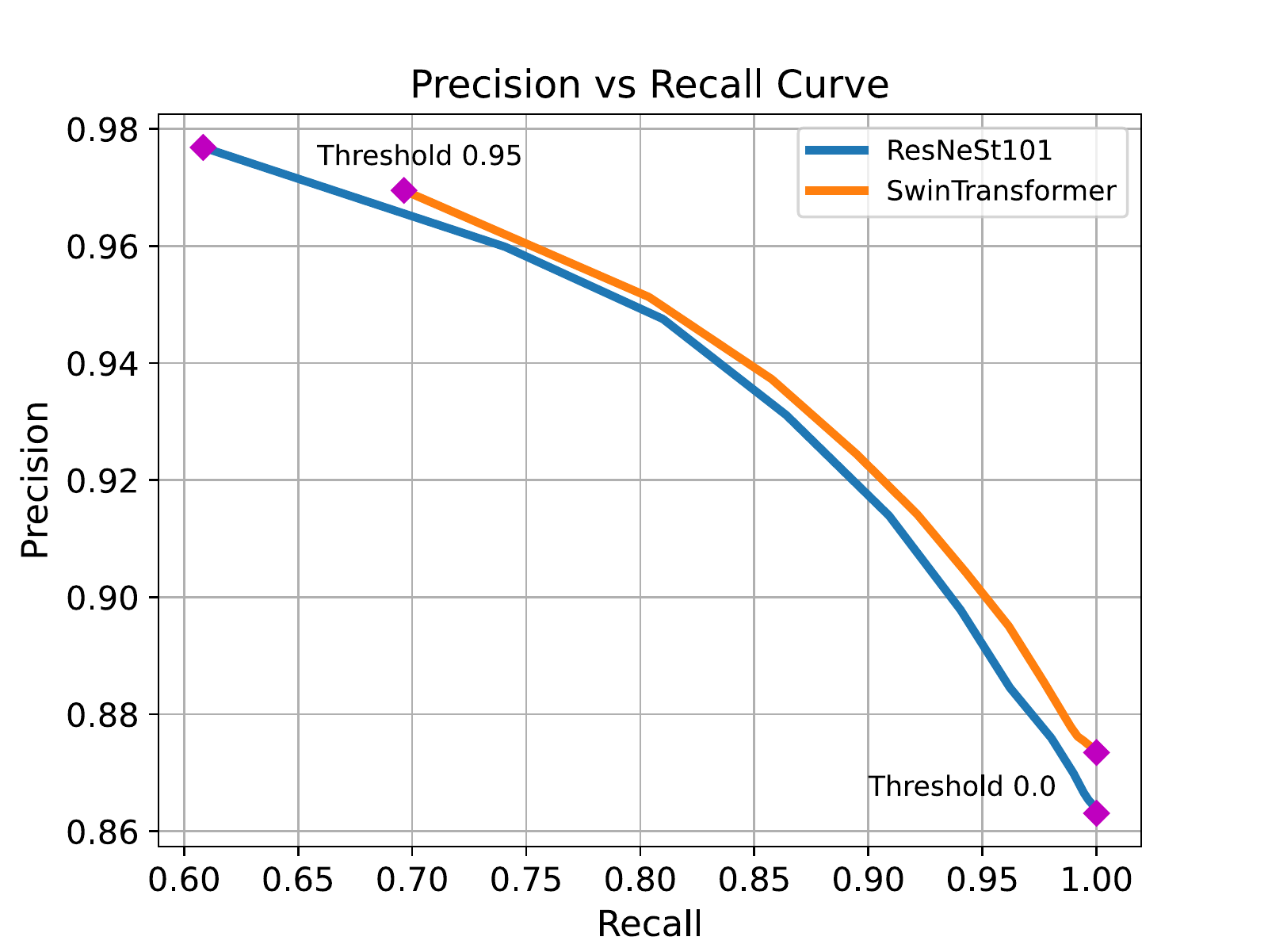}
	\caption{Precision-recall curve for the tested models.\label{fig:pr_curve}}
\end{figure}

The global precision-recall curve for all classes considered is shown in Figure~\ref{fig:pr_curve}. 

The graph confirms the good results, also represented by the other statistics presented earlier. It is close to the curve for the ideal classifier. It is also worth noting that in the curve for thresholds 0.95 (upper left corner), the sensitivity is high (0.6), which testifies to the model's reliable yet correct predictions. The high quality of the models is also confirmed by the high value of precision for the whole recall range.

\begin{figure}[!hp]
\centering
	\includegraphics[width=0.7\textwidth]{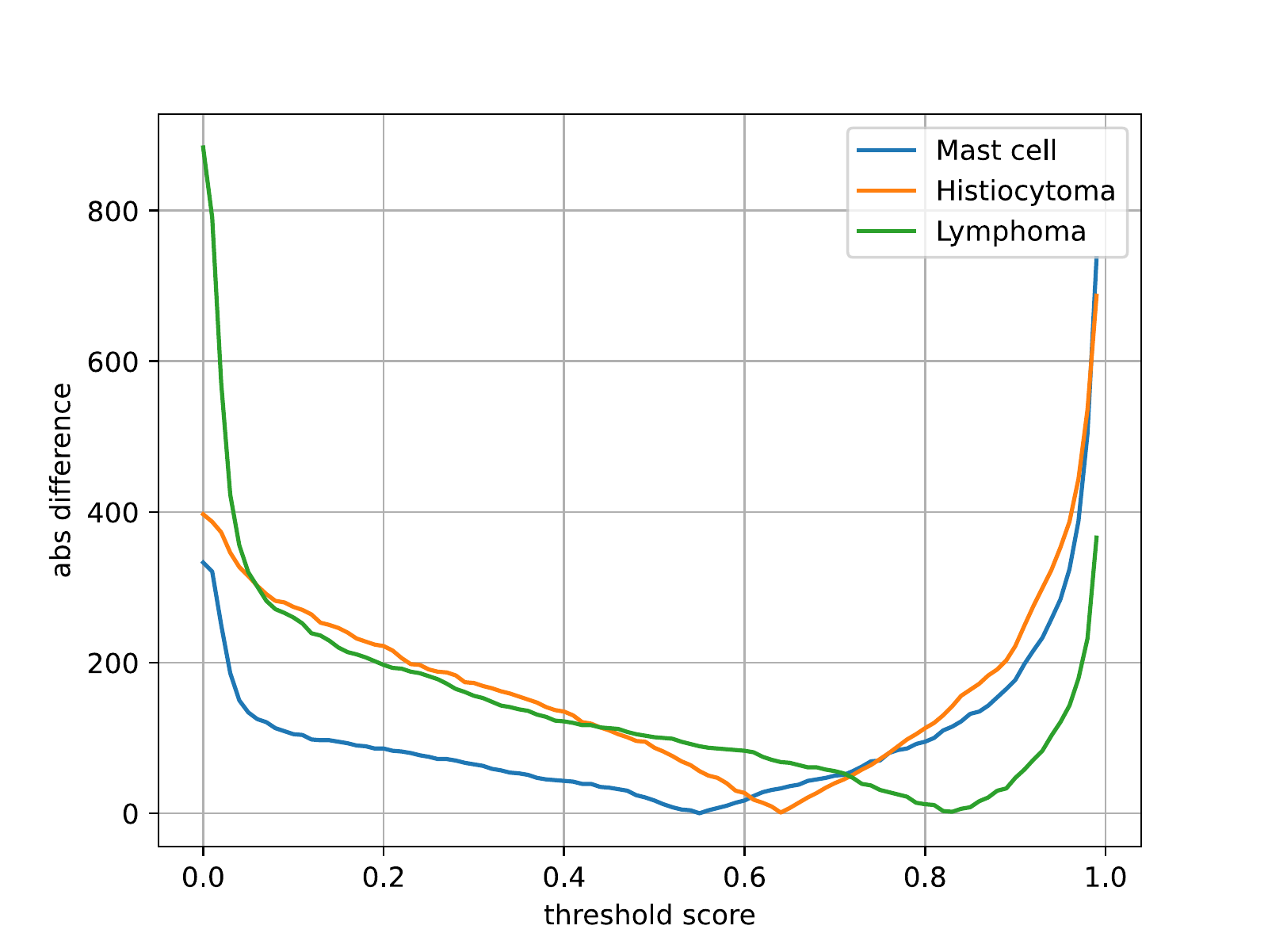}
	\caption{Absolute count difference of tumor cells instances vs. score threshold for tumor cells only. \label{fig:abs_diff_tumors}}
\end{figure} 

The additional visualization of the model performance was introduced in Figure \ref{fig:abs_diff_tumors}. It shows the absolute difference between the model prediction and the ground truth regarding the absolute number of cells. It can be noticed that there is a slight discrepancy between the tumors in terms of the optimal threshold value i.e. different threshold values are most favorable for different tumors. It ranges from approx. 0.55 to 0.85.  Nevertheless, the curves are still relatively flat i.e. the middle part is quite broad.

\begin{figure}[!hp]
\centering
	\includegraphics[width=0.75\textwidth]{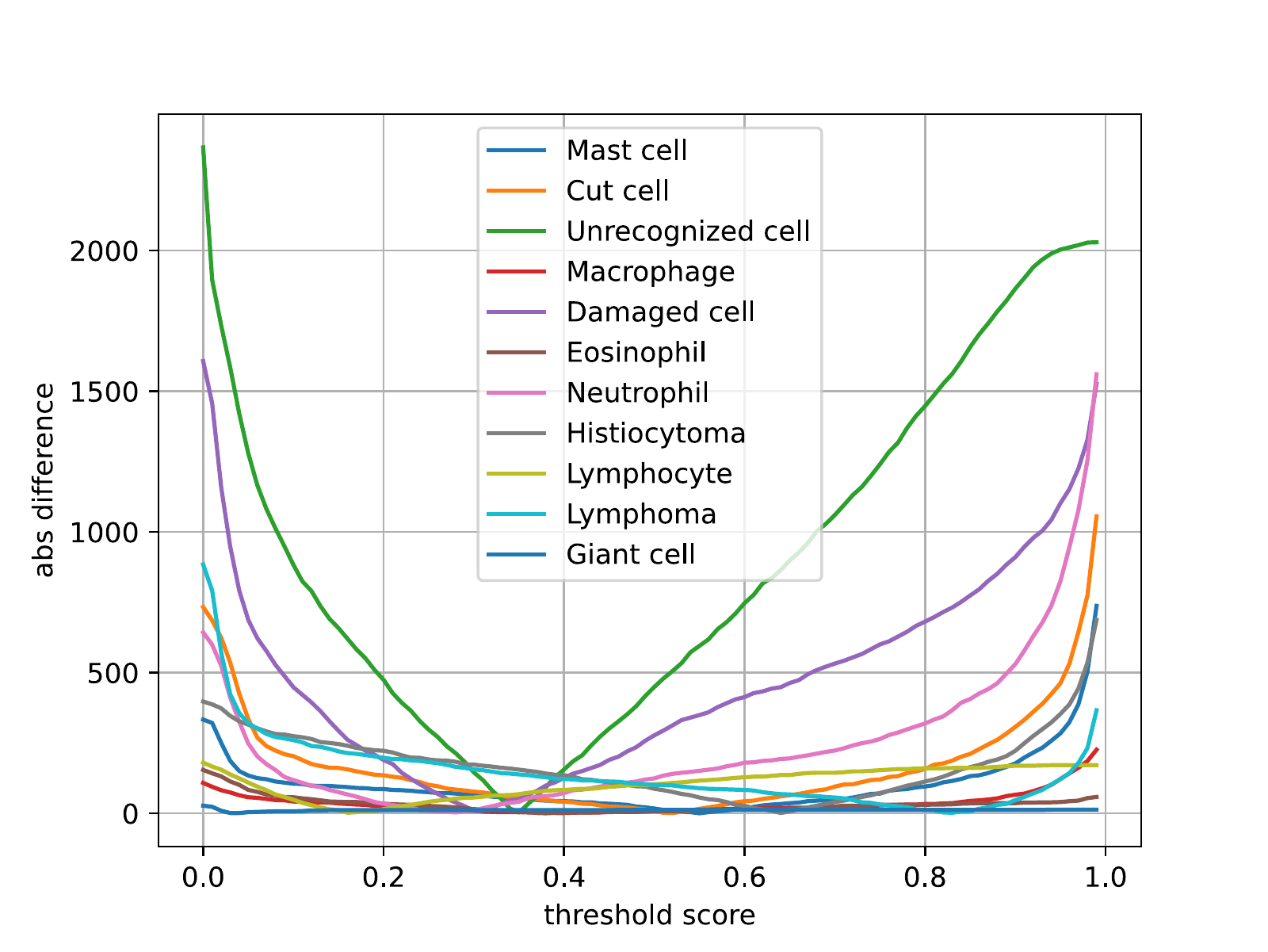}
	\caption{Absolute count difference of cell instances vs. score threshold for all cell types. \label{fig:abs_diff}}
\end{figure}

This observation is no longer valid when it comes to the analysis of all categories as presented in Figure \ref{fig:abs_diff}. Instead, the curves are steep, with a particular case of the unrecognized cell category, which takes a v-like shape. Such a constellation of figures results in a challenge when it comes to calibrating the model for the best performance.

\newpage

\subsection{Visual inspection of the results} \label{cha:visual_analysis}

Quality assessment metrics are very useful tools however it is also important to analyze selected results of the model outputs. For example, suppose the model assigned a cell to a category incorrectly. In that case, it is counted as a mistake, regardless of whether the classification can be easily or difficultly done by a human expert. Therefore, we present a series of prediction results to show the strengths and weaknesses of our current solution.

\begin{figure}[ht]
\centering
	\includegraphics[width=0.75\textwidth]{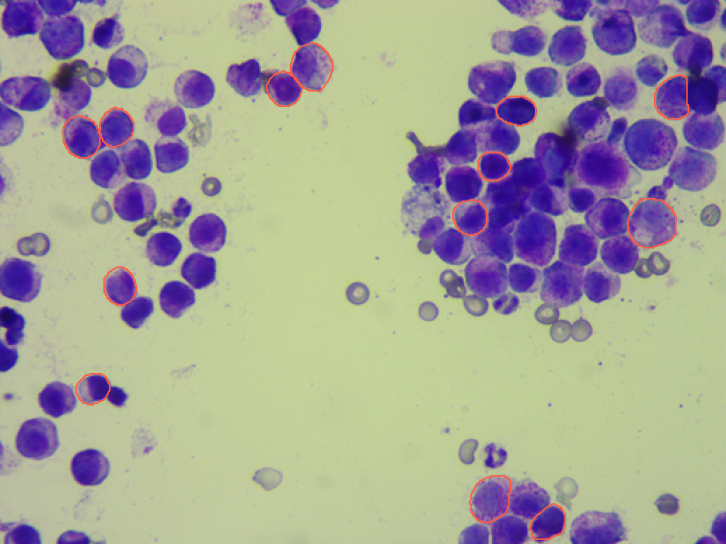}
 \\
	\includegraphics[height=3.6cm]{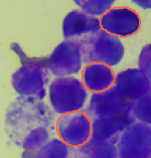}
	\includegraphics[height=3.6cm]{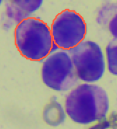}
	\caption{
	The example of incorrect classification of the unrecognized cell as a neoplastic mast cell. Zoomed fragments below.
	\label{fig:mistake:unrecognised-mast}}
\end{figure}

Analyzing the confusion matrices given in Section~\ref{cha:conf_mtx}, it is clear that most of the mistakes are within unrecognized, damaged, or cut cell classes. For example, in Figure \ref{fig:mistake:unrecognised-mast} model predicted selected cells as mastocytoma cells, while the correct label is unrecognized. Given that the photo was taken of the mastocytoma sample, the expert can tell they probably are mastocytoma cells; however, looking at single-cell only, they do not show specific features: they are very dark, and granulation is not apparent.

Performance metrics were analyzed with the IoU threshold set to 0.75 and the score threshold set to 0.75. They are reasonable values in the middle of the scale, giving good overall results. However, to explore the capabilities of the model throughout visual inspection it is interesting to set the threshold to extreme values. We will use score threshold values of 0.0 and 0.9 to highlight \textit{not-detected} and \textit{not-present} classes respectively. 

Model parameters used for this analysis are presented in Table \ref{tab:mistakes_swin}. It is worth adding that for \textit{not-present} setting, while the amount of misclassifications is smaller than in other settings, a significantly larger number of cells were not detected.

For each detected cell model gives probability, meaning how confident the model is that the cell is what it is, 0.95 is very high confidence and a mistake with that confidence is a serious error, 0.10 is low confidence and it is very probable that it is not correct so in general. We remove such low-confidence cells from the final results, the model returns quite a lot of them and most of them are not correct. It is sometimes the case that the cell is not detected because the model confidence is too low, like 0.60, in that case, we are generally on the right track and global improvements to the model should help (better backbone, more training samples). However, if we set the threshold to 0 and the model still did not detect the cell, it is more difficult to improve over that, hence the conclusion that they show model shortcomings.

\begin{table}[ht] 
\centering
 \caption{Comparison of CBNetV2 mistakes count for extreme and balanced threshold parameters \label{tab:mistakes_swin}}
	\begin{tabular}{|l|c|c|c|}
	\hline
	\textbf{Setup type}	& \textbf{IoU threshold}	& \textbf{Score threshold}	& \textbf{Mistakes count}\\ \hline
	
	Not-detected	& 0			    & 0 			& 844 	\\ \hline
	Not-present		& 0.5			& 0.9			& 239 	\\ \hline
	Balanced		& 0.75			& 0.75			& 677 	\\ \hline
 
	\end{tabular}
\end{table}

\newpage
\subsubsection{Not-detected setup}

To find cells that the model cannot detect, we set both threshold parameters to 0.0. This setting will highlight the most difficult cases for the model to find. 

\begin{figure}[!ht]
\centering
	\includegraphics[width=0.6\textwidth]{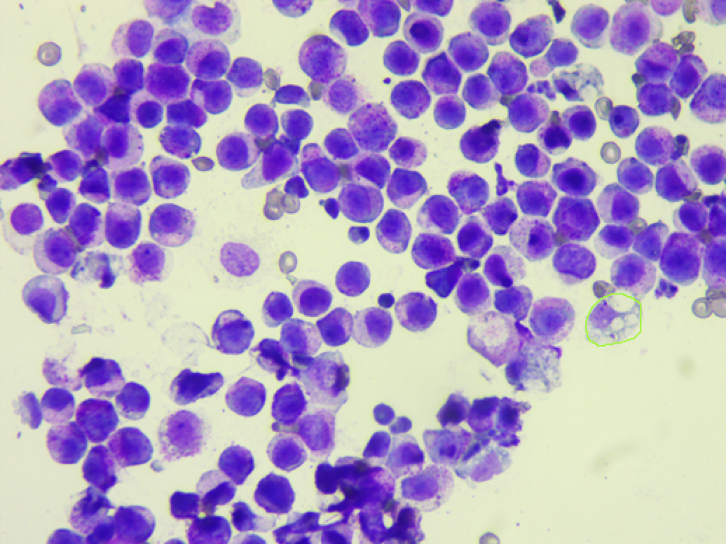}
	\includegraphics[height=3.2cm]{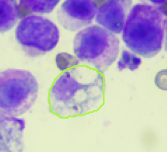}
	\caption{Undetected macrophage cell (in green). The zoomed fragment is below.}
 \label{fig:mistake:not_detected-Macrophage}
\end{figure}

The macrophage cell may look different depending on its current activity.  The ones that have not absorbed anything have uniform cytoplasm. However, the one in Figure~\ref{fig:mistake:not_detected-Macrophage} phagocytosed (i.e. 'ate' something), and it contains vacuoles with fragments of the absorbed material in its cytoplasm. Despite this, the macrophage that is visible in the picture (Fig.~\ref{fig:mistake:not_detected-Macrophage}) has a  preserved structure and should be recognized by the model. Unfortunately, it was not.

\begin{figure}[!ht]
\centering
	\includegraphics[width=0.65\textwidth]{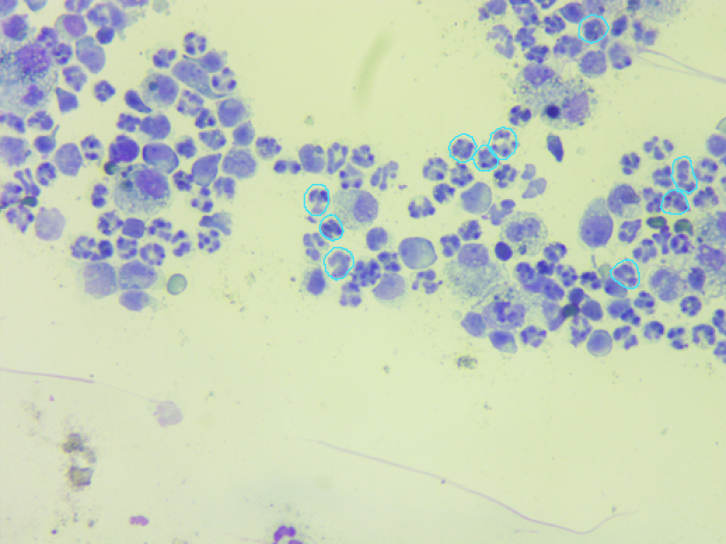} \\
	\includegraphics[height=3.6cm]{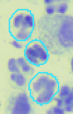}
	\includegraphics[height=3.6cm]{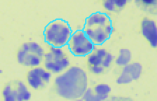}
	\caption{Undetected neutrophil cell (in blue). Zoomed fragments below.}
 \label{fig:mistake:not_detected-neutrophil}
\end{figure}

\newpage   
\hspace{10cm}

While for macrophages, difficulties in detection may be explained by their low number in the training dataset (Tab. \ref{tab:complete_results_with_count}), the neutrophil cells are pretty numerous. However, neutrophils found in picture slides are often in a state of disintegration and they are difficult to perceive. The example of unrecognised neutrophils are given in Figure \ref{fig:mistake:not_detected-neutrophil}.

\begin{figure}[ht]
\centering
	\includegraphics[width=0.7\textwidth]{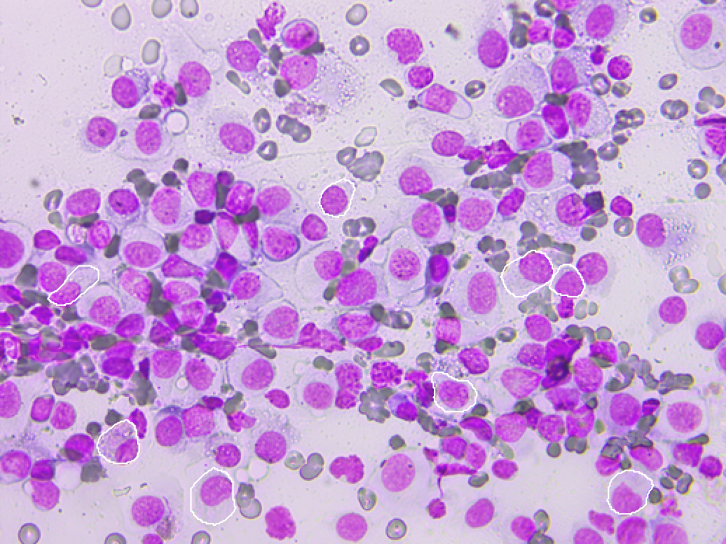} \\
	\includegraphics[height=2.8cm]{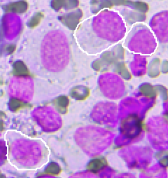}
	\includegraphics[height=2.8cm]{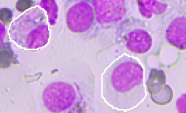}
	\caption{Undetected histiocytoma cell (surrounded with white borders). Zoomed fragments below.\label{fig:mistake:not_detected-histiocytoma}}
\end{figure} 

For the example shown in Figure~\ref{fig:mistake:not_detected-histiocytoma}, the cell has a preserved structure, the cell nucleus is slightly kidney-shaped, but the borders of the cytoplasm are very poorly visible. Therefore, it was hard for the model to detect the histiocytoma cell.

\begin{figure}[!ht]
\centering
	\includegraphics[width=0.73\textwidth]{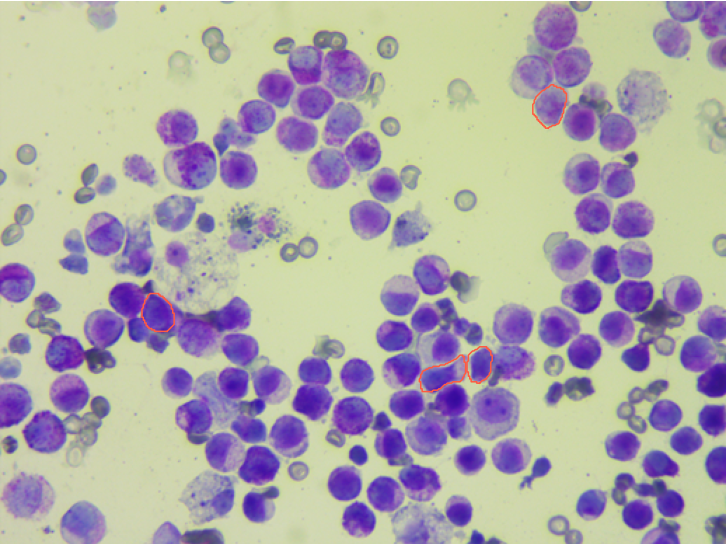} \\
	\includegraphics[height=3.5cm]{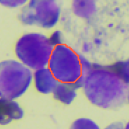}
	\includegraphics[height=3.5cm]{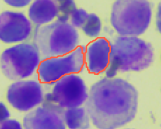}
	\caption{Mislabeled mastocytoma cell (surrounded with orange borders). Zoomed fragments below.\label{fig:mistake:not_detected-mastocytoma}}
\end{figure} 

Cells not detected by the model, highlighted in Figure~\ref{fig:mistake:not_detected-mastocytoma}, are most likely mastocytoma cells. It is a guess supported by the presence of other cells in the picture. The cells in question are distorted (not round), and no grain is visible in them. Their construction is obliterated. Labeling those cells as mastocytoma cells was an expert fault. In fact, they should be annotated as damaged or unrecognized.

\begin{figure}[!ht]
\centering
	\includegraphics[width=0.65\textwidth]{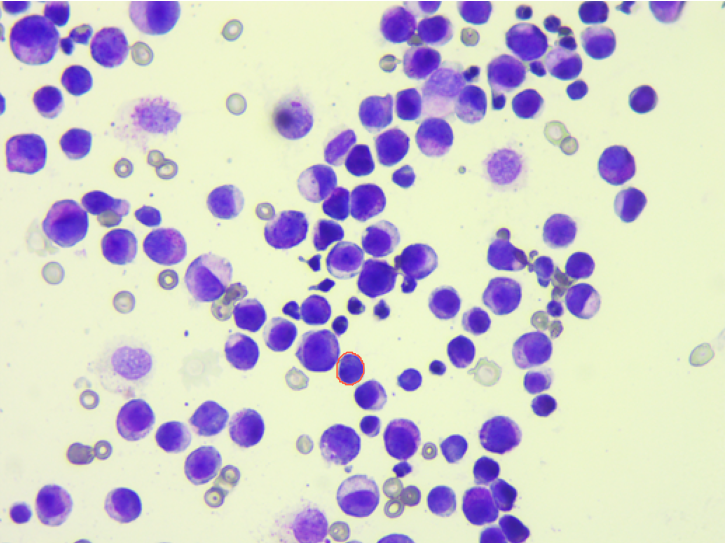}
	\includegraphics[height=3cm]{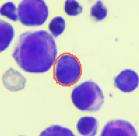}
	\caption{Difficult mastocytoma cell.
 \label{fig:mistake:not_present-mastocytoma}}
\end{figure}
\begin{figure}[!ht]
\centering
	\includegraphics[width=0.65\textwidth]{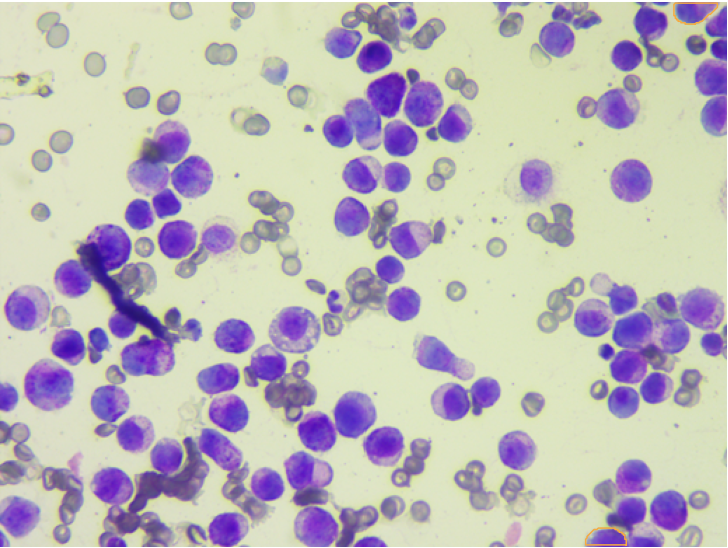} \\
	\includegraphics[height=2.5cm]{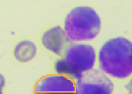}
	\includegraphics[height=2.5cm]{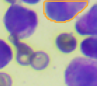}
	\caption{Annotation mistake - cut cell is present. Zoomed fragments below. \label{fig:mistake:not_present-cut}}
\end{figure}

 
Figure \ref{fig:mistake:not_present-cut} shows cut cell was correctly found. It was missed in the labeling process.

\newpage
\subsubsection{Not-present setup}

To analyze the \textit{not-present} faults, the model score threshold was set as high as 0.9. In this setup, we want to expose cases when cells are predicted by the model with high confidence but not present in the ground truth annotations. The value of the IoU threshold is less important for this analysis.

A phase of mastocytoma breakdown is captured in Figure~\ref{fig:mistake:not_present-mastocytoma}. The situation is ambiguous because the cell is actually decaying, but it can still be considered a mastocytoma cell.

Figure~\ref{fig:mistake:not_present-mastocytoma}. The situation is ambiguous because the cell is actually decaying, but it can still be considered a mastocytoma cell.

Visual inspection revealed that most of the \textit{not-present} faults resulted in classification dis-ambiguities that are unavoidable in the annotation process done by an expert. In most cases, the cells detected by the model were damaged, or unrecognized, or the classification was difficult for other reasons. This result shows the robustness of the model as even though it makes mistakes, it can still help improve human-made annotations.

The analysis provided in this section concentrated on the correctness of the class types assigned to the detected objects. It is worth noting that the instance segmentation performance also includes a flawless prediction of the borders of an object. In many presented cases of model errors, even if the category was mispredicted, the borders were close to the ground truth. It shows that the geometrical properties of the cells are less challenging to predict.

\subsection{Conclusions for dataset labeling} \label{cha:labelong_counclusions}
Labeling the dataset for instance segmentation task is challenging, especially when it requires finding borders between groups of closely coupled cells. On the one hand, marking separately every single cell is very time-consuming and error-prone; on the other hand, adopting an approach of marking a whole group of cells sometimes leads to skipping some vital elements of the image. Consequently, a dedicated annotation software was introduced, which fosters a labeling process by tentatively marking cells that the doctor later accepts. Introducing the software did not address all the mistakes in the visual data analysis step. However, it helped to increase labeling efficiency. 

It is worth noting that damaged cells are especially challenging to label. For example, it may lead an expert to a dilemma if a given cell is cancerous and should be classified as a tumor category. Neutrophils are challenging to detect and recognize because cytoplasm is not detectable with routine stainings. The plasmatic membrane is barely perceptible. Furthermore, neutrophils have nuclei with multiple, usually 3-5 lobes connected by thin chromatin striae. The exact shape of the neutrophil's nucleus depended on the cell age, presence of bacteria, and damage introduced during the preparation of the slides.

\section{Discussion}

Both of the presented models allow the veterinary system to detect and recognize  selected categories of cells effectively. The images processed by the models were similar to those annotated by the experts, and high metrics scores confirmed it.

The SwinTransformer-based algorithm performs noticeably better. This finding can be confirmed in the mAP results and the confusion matrices. The SwinTransformer concept is much newer than ResNeSt101, and it also produces better results on the MS COCO. We have selected models based on MS COCO results, hence the overall improvement.

There are some errors, but their percentage in the total volume of cells is relatively small.  Additionally, the errors primarily concern cells whose number in the training set was smaller than the cancer cells. Therefore, discrepancies are entirely ignorable when general diagnosis must be inherited from cell classification.

Most importantly, the models' detections are of high quality - the models are confident in their good decisions, and in addition, they have high IoU values. This outcome is visible in the mAP metrics for the high IoU thresholds and the error matrices plotted for their thresholds.

\section{Conclusions and future work}
The paper addresses the demanding tasks of medical image segmentation. The challenge of the process mainly stems from the data ambiguity and the labeling process. In many cases, cells presented in images are damaged and hard to classify into one of the tumor categories.
It is worth underlining that the model did not make a single mistake in classifying one of the tumor cells to a different tumor cell type.

We proposed the training pipeline, which is well suited for the task, and also presented details of its configuration. Furthermore, a detailed analysis of the mistakes made by the model was conducted. It shows that there are some cases that may be easy for the model and hard for the vet doctor to spot. This results from multiple reasons, but it's vital to know what types of cancers and images are prone to that kind of challenge, as we presented in the paper.

In future work, we are going to extend the system with more types of cells and also include inflammation. A very interesting direction of the research is building a system capable of combining the detection of inflammation and cancerous changes to infer more challenging patient cases.

\appendix
\newpage
\section{Extended metrics}

\label{appendix_A}

\begin{table}[hp]
\centering
\caption{Segmentation metrics calculated for all categories of the test set\label{tab:tumor_cells_metrics_extended}}
\begin{tabular}{|c|c|c|c|c|}
\hline
\textbf{Category}             & \textbf{Metric}   & \textbf{IoU threshold} & \textbf{ResNeSt101} & \textbf{CBNetV2} \\ \hline
\multirow{4}{*}{Mastocytoma}  & Average Precision & 0.50:0.95              & \textbf{0.769}      & 0.764            \\ \cline{2-5} 
                              & Average Precision & 0.50                   & \textbf{0.940}      & \textbf{0.940}   \\ \cline{2-5} 
                              & Average Precision & 0.75                   & \textbf{0.940}      & 0.930            \\ \cline{2-5} 
                              & Average Recall    & 0.50:0.95              & 0.806               & 0.799            \\ \hline
\multirow{4}{*}{Histiocytoma} & Average Precision & 0.50:0.95              & 0.665               & \textbf{0.671}   \\ \cline{2-5} 
                              & Average Precision & 0.50                   & \textbf{0.877}      & 0.868            \\ \cline{2-5} 
                              & Average Precision & 0.75                   & 0.838               & \textbf{0.843}   \\ \cline{2-5} 
                              & Average Recall    & 0.50:0.95              & 0.740               & \textbf{0.745}   \\ \hline
\multirow{4}{*}{Lymphoma}     & Average Precision & 0.50:0.95              & 0.743               & \textbf{0.760}   \\ \cline{2-5} 
                              & Average Precision & 0.50                   & 0.955               & \textbf{0.964}   \\ \cline{2-5} 
                              & Average Precision & 0.75                   & 0.946               & \textbf{0.956}   \\ \cline{2-5} 
                              & Average Recall    & 0.50:0.95              & 0.796               & \textbf{0.812}   \\ \hline
\multirow{4}{*}{Cut}          & Average Precision & 0.50:0.95              & \textbf{0.521}      & 0.520            \\ \cline{2-5} 
                              & Average Precision & 0.50                   & 0.822               & \textbf{0.833}   \\ \cline{2-5} 
                              & Average Precision & 0.75                   & \textbf{0.636}      & 0.630            \\ \cline{2-5} 
                              & Average Recall    & 0.50:0.95              & 0.595               & \textbf{0.611}   \\ \hline
\multirow{4}{*}{Damaged}      & Average Precision & 0.50:0.95              & 0.430               & \textbf{0.441}   \\ \cline{2-5} 
                              & Average Precision & 0.50                   & 0.639               & \textbf{0.649}   \\ \cline{2-5} 
                              & Average Precision & 0.75                   & 0.547               & \textbf{0.557}   \\ \cline{2-5} 
                              & Average Recall    & 0.50:0.95              & 0.526               & \textbf{0.540}   \\ \hline
\multirow{4}{*}{Unrecognized} & Average Precision & 0.50:0.95              & 0.291               & \textbf{0.303}   \\ \cline{2-5} 
                              & Average Precision & 0.50                   & 0.444               & \textbf{0.454}   \\ \cline{2-5} 
                              & Average Precision & 0.75                   & 0.365               & \textbf{0.385}   \\ \cline{2-5} 
                              & Average Recall    & 0.50:0.95              & 0.495               & \textbf{0.513}   \\ \hline
\multirow{4}{*}{Macrophage}   & Average Precision & 0.50:0.95              & 0.498               & \textbf{0.531}   \\ \cline{2-5} 
                              & Average Precision & 0.50                   & 0.653               & \textbf{0.689}   \\ \cline{2-5} 
                              & Average Precision & 0.75                   & 0.628               & \textbf{0.667}   \\ \cline{2-5} 
                              & Average Recall    & 0.50:0.95              & 0.579               & \textbf{0.624}   \\ \hline
\multirow{4}{*}{Eosinophil}   & Average Precision & 0.50:0.95              & 0.251               & \textbf{0.515}   \\ \cline{2-5} 
                              & Average Precision & 0.50                   & 0.328               & \textbf{0.673}   \\ \cline{2-5} 
                              & Average Precision & 0.75                   & 0.328               & \textbf{0.662}   \\ \cline{2-5} 
                              & Average Recall    & 0.50:0.95              & 0.447               & \textbf{0.619}   \\ \hline
\multirow{4}{*}{Neutrophil}   & Average Precision & 0.50:0.95              & 0.568               & \textbf{0.581}   \\ \cline{2-5} 
                              & Average Precision & 0.50                   & 0.809               & \textbf{0.818}   \\ \cline{2-5} 
                              & Average Precision & 0.75                   & 0.750               & \textbf{0.779}   \\ \cline{2-5} 
                              & Average Recall    & 0.50:0.95              & 0.640               & \textbf{0.652}   \\ \hline
\multirow{4}{*}{Lymphocyte}   & Average Precision & 0.50:0.95              & 0.226               & \textbf{0.316}   \\ \cline{2-5} 
                              & Average Precision & 0.50                   & 0.330               & \textbf{0.473}   \\ \cline{2-5} 
                              & Average Precision & 0.75                   & 0.291               & \textbf{0.435}   \\ \cline{2-5} 
                              & Average Recall    & 0.50:0.95              & 0.373               & \textbf{0.477}   \\ \hline
\multirow{4}{*}{Giant cell}   & Average Precision & 0.50:0.95              & 0.0                 & \textbf{0.205}   \\ \cline{2-5} 
                              & Average Precision & 0.50                   & 0.0                 & \textbf{0.228}   \\ \cline{2-5} 
                              & Average Precision & 0.75                   & 0.0                 & \textbf{0.228}   \\ \cline{2-5} 
                              & Average Recall    & 0.50:0.95              & 0.0                 & \textbf{0.257}   \\ \hline
\end{tabular}
\end{table}

\begin{figure}[!ht]
	\includegraphics[width=\textwidth]{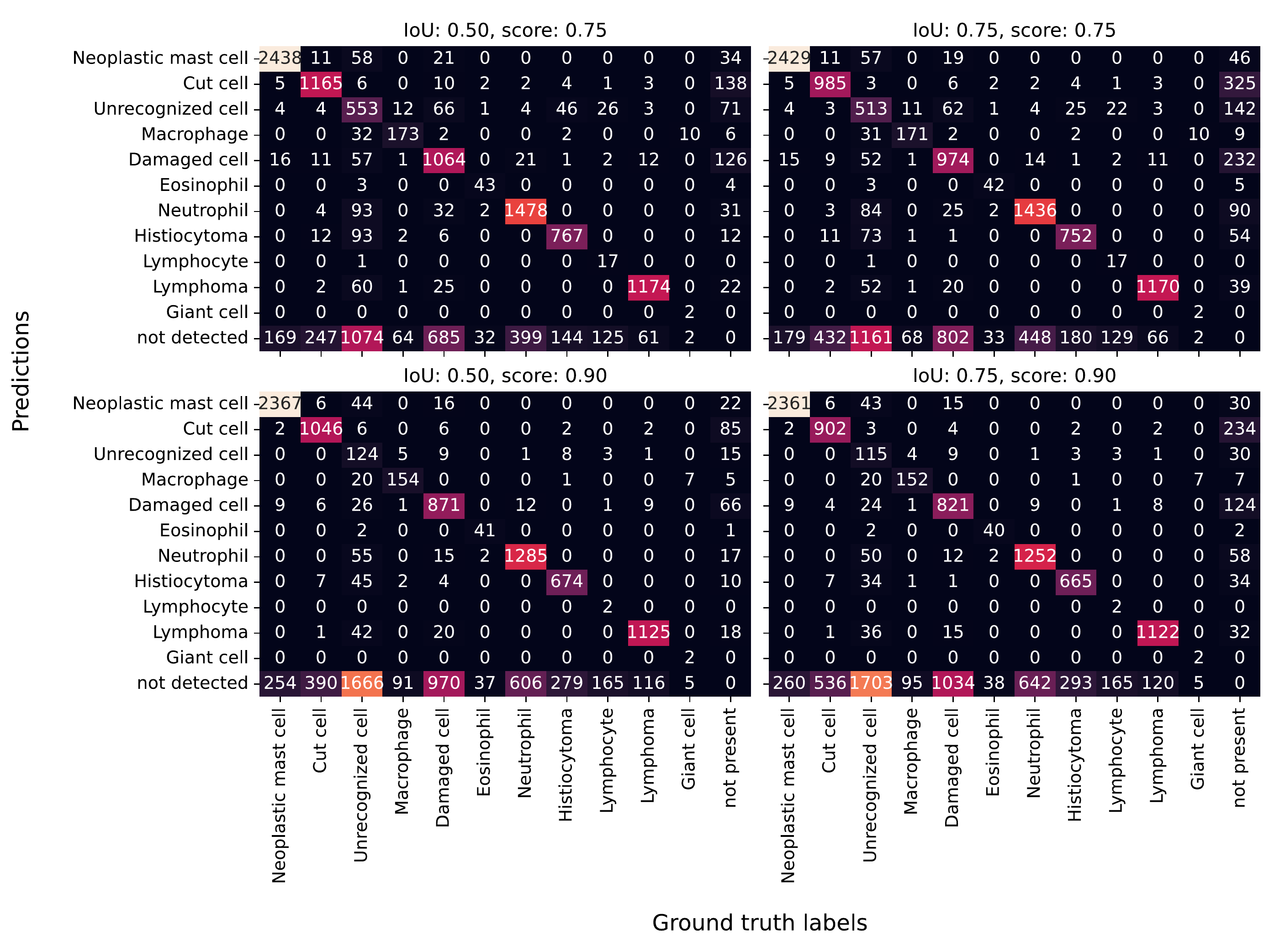}
	\caption{Extended confusion matrices for CBNetV2 model.
	\label{fig:swin_extended_cms}}
\end{figure} 

\newpage

\begin{figure}[!ht]
    \centering
	\includegraphics[width=\textwidth]{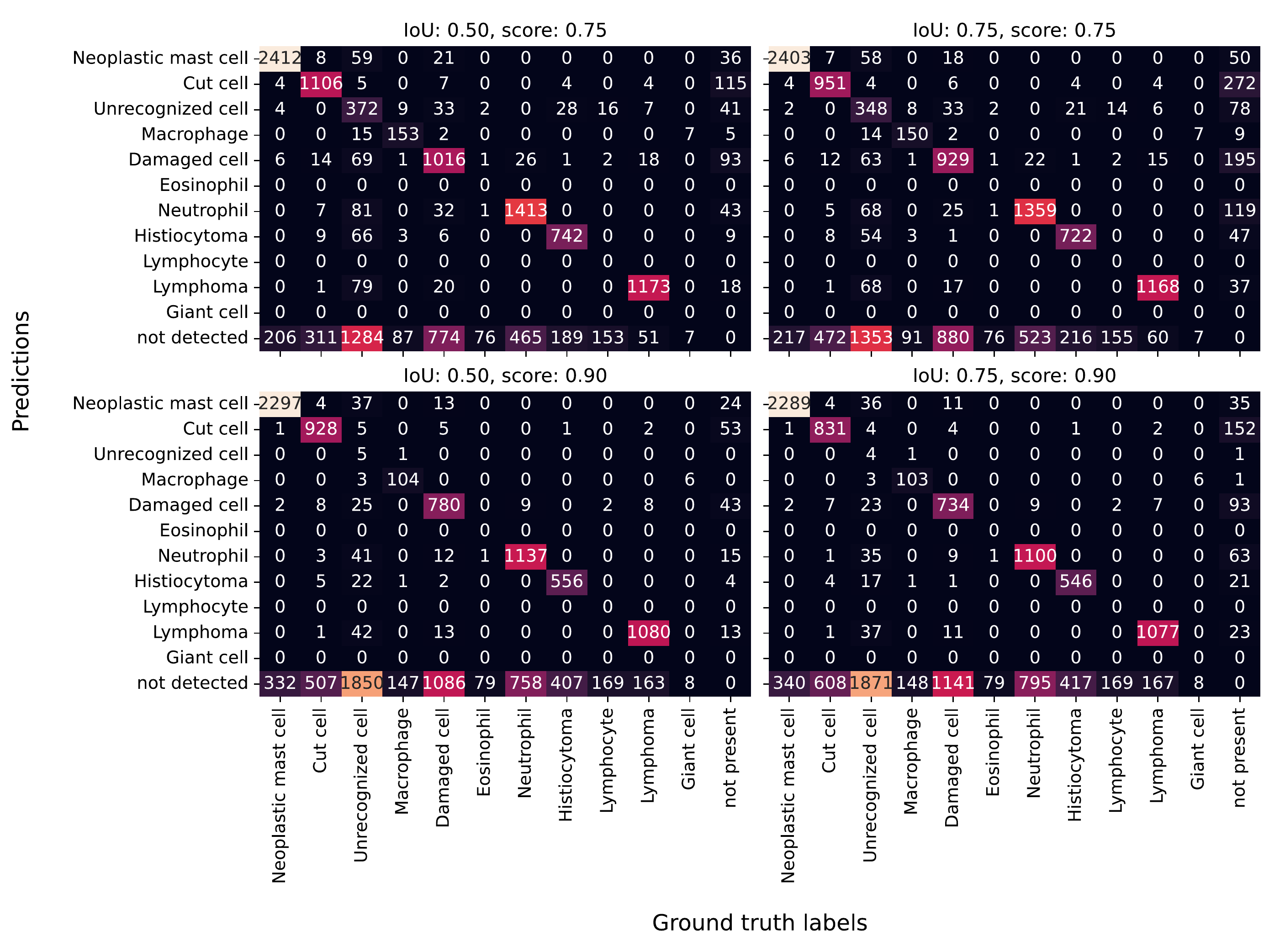}
 	\caption{Extended confusion matrices for ResNeSt101 model.
 	\label{fig:resnest_extended_cms}}
\end{figure} 

\hspace{10cm}

\pagebreak
\bibliographystyle{unsrt}  
\bibliography{paper}

\end{document}